\documentclass[runningheads]{llncs}
\usepackage{graphicx}
\usepackage{tablefootnote}
\usepackage{multirow}
\usepackage{amsfonts}
\usepackage{url}
\usepackage{hyperref}
\usepackage{subcaption}

\begin{document}
\title{{\em An EcoSage Assistant:} Towards Building A Multimodal Plant Care Dialogue
Assistant}

\author{Mohit Tomar\inst{1} \and   
Abhisek Tiwari\inst{1} \and
Tulika Saha\inst{2} \and Prince Jha\inst{1} \and Sriparna Saha\inst{1}}
\institute{Department of Computer Science and Engineering \\ Indian Institute of Technology Patna, Patna, India \and Department of Computer Science and Engineering, \\University of Liverpool, United Kingdom \\
\email{\{mohitsinghtomar9797, abhisektiwari2014, sahatulika15, jhapks1999, sriparna.saha\}@gmail.com }}
\titlerunning{Towards Building A Multimodal Plant Care Dialogue
Assistant}

\maketitle              % typeset the header of the contribution
\vspace{-1.5em}
\begin{abstract}
% In the midst of an increasingly urbanized world, where concrete landscapes dominate, and nature seems to recede, the concept of nurturing green life has taken on a new dimension of importance.
In recent times, there has been an increasing awareness about imminent environmental challenges, resulting in people showing a stronger dedication to taking care of the environment and nurturing green life. The current \$19.6 billion indoor gardening industry, reflective of this growing sentiment, not only signifies a monetary value but also speaks of a profound human desire to reconnect with the natural world. However, several recent surveys cast a revealing light on the fate of plants within our care, with more than half succumbing primarily due to the silent menace of improper care. Thus, the need for accessible expertise capable of assisting and guiding individuals through the intricacies of plant care has become paramount more than ever. In this work, we make the very first attempt at building a plant care assistant, which aims to assist people with plant(-ing) concerns through conversations. We propose a plant care conversational dataset named {\em Plantational}, which contains around 1K dialogues between users and plant care experts. Our end-to-end proposed approach is two-fold : (i) We first benchmark the dataset with the help of various large language models (LLMs) and visual language model (VLM) by studying the impact of instruction tuning (zero-shot and few-shot prompting) and fine-tuning techniques on this task; (ii) finally, we build \textit{EcoSage}, a multi-modal plant care assisting dialogue generation framework, incorporating an adapter-based modality infusion using a gated mechanism. We performed an extensive examination (both automated and manual evaluation) of the performance exhibited by various LLMs and VLM in the generation of the domain-specific dialogue responses to underscore the respective strengths and weaknesses of these diverse models\footnote{The dataset and code are available at \url{https://github.com/mohit2b/EcoSage}}. 

\keywords{Plant Care  \and Virtual Assistant \and Large Language Models (LLMs) \and Multi-modal infusion \and Dialogue Generation}
\end{abstract}
\section{Introduction}

% Agriculture has been a vital endeavour since the beginning of human civilization, with reliance extending from providing oxygen to sustaining nutrition. 
According to a recent survey conducted by the International Labour Organization (ILO), over a quarter of the global population continues to depend solely on one profession, which is agriculture\footnote{\url{https://data.worldbank.org/indicator/SL.AGR.EMPL.ZS}}. Beyond financial advantages, the environmental benefits derived from agriculture and plantation practices are substantial. In the near future, imminent environmental challenges like climate change, biodiversity loss, and soil degradation are expected to escalate. Plantation holds promise as a potent solution to these issues. By strategically planting diverse tree species, particularly those with substantial carbon sequestration capabilities, one can mitigate climate change.
% For example, according to information from the Food and Agriculture Organization (FAO), a sole hectare of rapidly growing eucalyptus plantation can capture as much as 25 tons of carbon dioxide annually, highlighting the significant contribution to carbon sequestration \cite{chavan2023carbon}. 
Over the last decade, a deepened awareness has led individuals toward a growing emphasis on plantation endeavours, with the indoor planting market projected to attain a size of 31 billion by 2032, with a Compound Annual Growth Rate (CAGR) of 4.8\%.
\vspace{-1em}

\begin{figure}[ht]
    \centering
    \includegraphics[width=\linewidth]{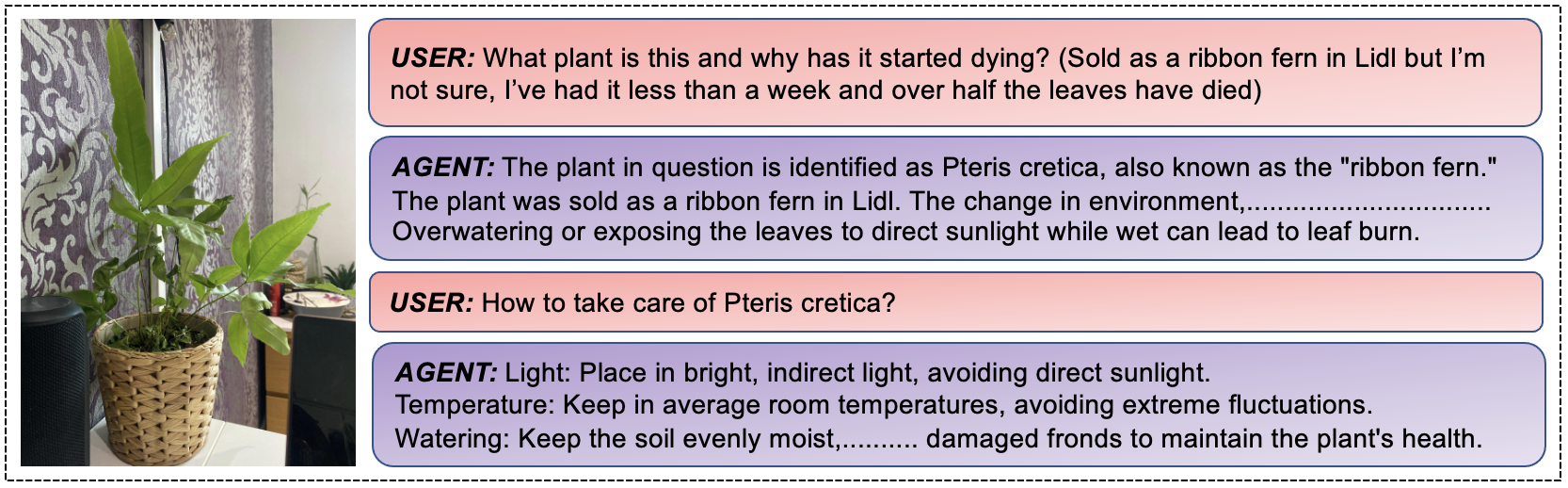}
  \vspace{-1.5em}
    \caption{A conversational illustration between a user and the agent}
     \vspace{-1em}
    \label{dialog_example}
\end{figure}

\vspace{-0.5em}
The rate at which people dedicate efforts to planting doesn't appear to match its potential growth. This is primarily attributed to inadequate management and a lack of guidance concerning plant(ing)-related issues. Individuals are quite reluctant to follow up with plant specialists. However, with the advancement of AI-based agricultural assistants and web technology, we often look for solutions to different issues over the internet. People often post their issues over some popular discussion forums like \textit{Reddit} and \textit{Houzz} with dedicated threads such as \textit{Houseplants}\footnote{\url{https://www.reddit.com/r/houseplants/}} and \textit{House-plants}\footnote{\url{https://www.houzz.com/discussions/house-plants}} comprising of more than 20K posts and 1.5M registered users. The primary aim of these communities revolves around engaging in discussions and promoting the care and welfare of indoor plants. Nonetheless, a single, irrelevant recommendation can potentially exacerbate a plant's condition, and the assurance of real-time responses in such forums is not guaranteed. To assist such online plant healthcare seekers, we build a plant care assistant that can assist people with plant(-ing)-related issues and motivate them for specialist consultations whenever needed. An example of such a conversation between a plant assistant and a user is illustrated in Figure \ref{dialog_example}.

% In the realm of natural language processing, distinct phases have marked the progression of text processing capabilities in models:  seq2seq \cite{sutskever2014sequence,bahdanau2014neural}, transformer \cite{vaswani2017attention}, and currently, the emergence of large language models (LLMs). 
The evolution of Large Language Models (LLMs) and ChatGPT has made the general audience believe that all-natural language processing problems have been solved. However, many significant challenges persist, particularly within domains constrained by limited resources. Our preliminary assessments of LLMs for plant care assistance reveal a substantial gap between anticipated outcomes and those generated. We also delve into the visual language model, which offers an improvement over previous models \cite{zhu2023minigpt} but still falls short of human expectations. Motivated by the burgeoning interest in plant care and driven by these limitations, we make the first move to investigate some fundamental research questions related to plant assistance response generation and build an assistant called {\em EcoSage} to provide initial guidance to plant(-ing)-related issues. The {\em EcoSage} Assistant is meant to seek user issues and pose further queries to gain a comprehensive understanding of the issue and subsequently, provide suggestions and responses to assist the support seekers. Confronted by the scarcity of conversational data in plant care, we undertook the initiative to create a plant care conversational dataset named {\em Plantational} encompassing a range of plant(-ing)-related issues. We further enrich the dataset by assigning intent and dialogue act (DA) categories to each dialogue and utterances within them.

\hspace{-0.55cm}\textbf{Research Questions.} In this work, we aim to answer the following three research questions: \textbf{(i)} Can existing state-of-the-art LLMs adequately offer initial recommendations related to plant(-ing)-related queries? \textbf{(ii)} Do LLMs that take images into account comprehend concerns more effectively and produce suitable and better responses? \textbf{(iii)} What is the appropriate way to gauge the effectiveness of the response generation model? Are metrics based on n-gram overlap sufficient for the evaluation, and are they consistent with semantic evaluation? 

\hspace{-0.55cm}\textbf{Key Contributions.} The key contributions of the work are as follows : 

\begin{itemize}
\vspace{-0.5em}
    \item We first build {\em Plantational}, a multi-modal, multi-turn plant care conversational dataset, which consists of around 1K conversations spanning over 4900 utterances. The dataset is the first plant-based conversational corpus containing plant-related discussions intended to aid users in the plantation. 

    \item The work investigates the efficacy of different LLMs and VLM for plant care assistants in both zero-shot and few-shot settings. 

    \item Motivated by the need for a proficient plant care assistant, we build a plant care response generation model incorporating an adapter-based modality infusion and fine-tuning mechanisms.
    
    \item The proposed model outperforms all baselines in zero-shot and few-shot settings across almost every evaluation metric by a significant margin. 
\end{itemize}

\section{Related Works}
 The current work is mainly related to the following three research areas: Dialogue generation, Large language models (LLMs), and Visual language models (VLMs). The following paragraphs summarize the relevant works. 

\hspace{-0.53cm}\textbf{Dialogue Generation.} 
DialoGPT \cite{zhang2019dialogpt} trains a conversation generation model based on Reddit comments. GODEL \cite{peng2022godel} uses grounded pre-training on external text for effectively generating a response to a conversation. Recently, reinforcement learning from human feedback is also utilized to train models \cite{bai2022training,glaese2022improving,schulman2022chatgpt} to train dialogue agents to be helpful and harmless.

\hspace{-0.53cm}\textbf{Large Language Models.} (LLM) has shown great results in various NLP tasks. Their progress started with models such as BERT \cite{devlin2018bert}, GPT \cite{radford2018improving}, and T5 \cite{raffel2020exploring}. They are often trained in mask language modeling or next token prediction objectives on a large chunk of the internet. GPT-3 \cite{brown2020language}, a 175 billion parameters model, achieved breakthroughs on many language tasks. This resulted in the development of more models, such as Gopher \cite{rae2021scaling}, OPT \cite{zhang2022opt}, Megatron-Turing NLG \cite{smith2022using}, PaLM \cite{chowdhery2022palm}, LLaMA \cite{touvron2023llama}. Also, fine-tuning LLM on instructions, FLAN \cite{chung2022scaling} and human feedback InstructGPT \cite{ouyang2022training} have achieved great results. 

\hspace{-0.53cm}\textbf{Visual Language Models.} Recently, large language models have been used as decoders for solving multimodal tasks. Flamingo \cite{alayrac2022flamingo} utilizes a frozen vision encoder and language decoder and trains only the perceiver resampler and cross-attention layers to achieve impressive few-shot performance on vision-language tasks. GPT-4 \cite{OpenAI2023GPT4TR} takes text and image as input and produces text as output. It achieves state-of-the-art performance on a range of NLP tasks. Open-source multimodal models \cite{anas_awadalla_2023_7733589,zhu2023minigpt,liu2023visual} have also been released that insert only a few trainable layers such as cross attention layer or linear projection layer to align visual features with textual features.

\hspace{-0.53cm}\textbf{Dialogue Systems for Social Good.} There have been multitude of works focused on developing conversational AI systems based on the social good theme serving various social good goals such as healthcare \cite{tiwari2022symptoms} including mental health \cite{saha2022towards,saha2022shoulder,saha2022mental,saha2021large}, education \cite{jain2023can,jain2023t,jain2022domain} etc.

% \textbf{In Context Learning}

% \textbf{Parameter Efficient Fine Tuning}

\section{Dataset}
Motivated by the unavailability of any conversational plant care assistance dataset and its significance in the present context, we endeavoured to curate a conversational plant care assistance corpus, {\em Plantational}. In this section, we discuss the details of the dataset and its curation.

\subsection{Data Collection}

\begin{figure*}[ht!]
    \centering
    \includegraphics[width=0.9\linewidth]{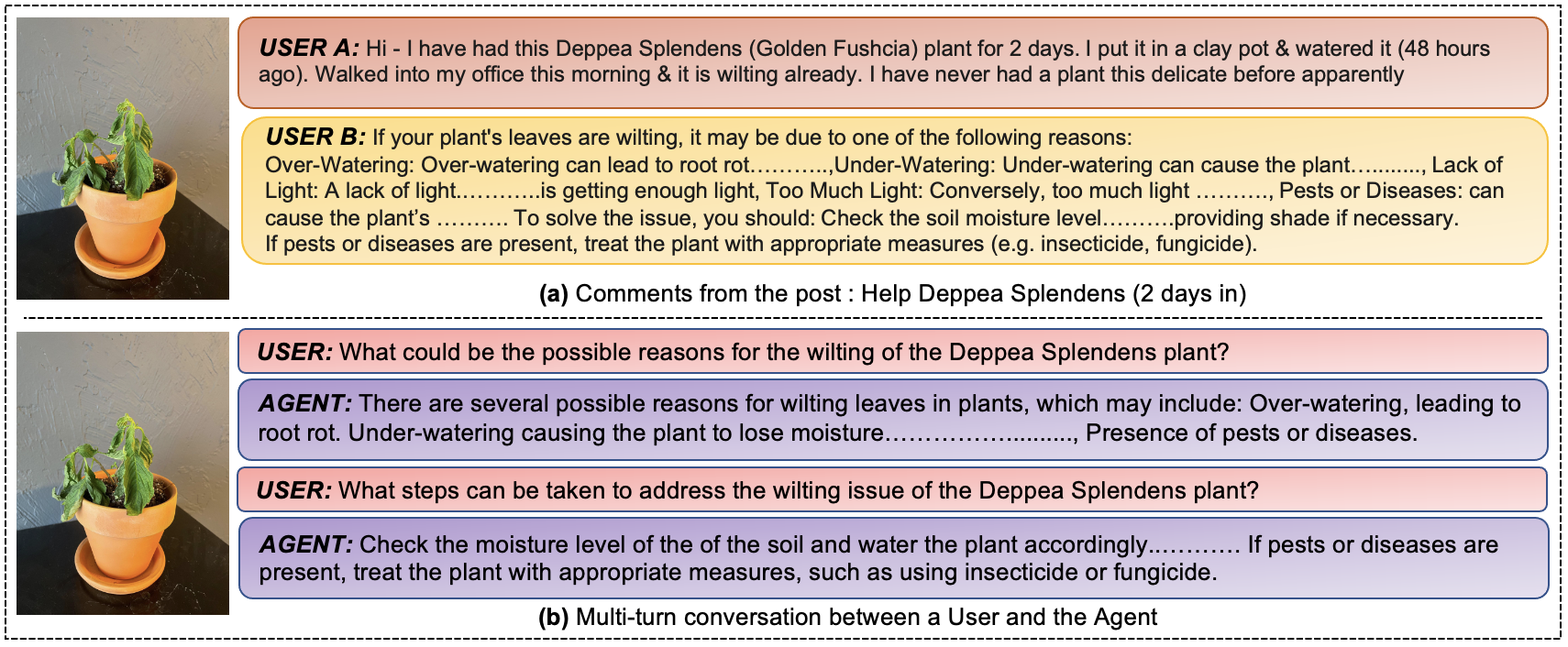}
    \setlength{\abovecaptionskip}{1pt}
  \setlength{\belowcaptionskip}{-5pt}
    \caption{(a) Indicates original Reddit post; (b) Converted Reddit post into a conversation between a User and the Agent}
    \vspace{-1em}
    \label{dialog_creation}
\end{figure*}

We initially conducted a thorough survey of datasets about plants and noted that all the existing research studies \cite{singh2020plantdoc,fenu2021diamos,liu2021plant} are primarily concentrated on classifying plant diseases based on images. The investigation revealed an absence of any dialogic dataset related to plants, whether in textual form or encompassing multimodal elements. However, we identified some potential discussion sites where people seek support to various problems relating to plants. The two most well-known of these forums are \textit{Reddit} and \textit{Houzz}. We found several dedicated sub-threads in these forums, such as \textit{house-plants}, with more than a million registered users and over twenty thousand postings. Each time a user posted about a plant(-ing)-related issue, other platform users tried offering advice and suggestions. One such example of a conversation from these forums is shown in Figure \ref{dialog_creation}a. 

In collaboration with two domain experts, we initiated an assessment of query quality and comment credibility for 50 posts extracted from the \textit{house-plants} forum. Our observations yielded the following insights: (i) Users adeptly articulated their queries, often accompanied by relevant images; (ii) While not all comments exhibited high credibility, those with substantial upvotes notably contributed to the discourse; (iii) Robust user engagement was evident as users actively participated in discussions by responding to other user's comments. Given the abundance of high-quality posts within sufficiently extensive forums, we selected 1150 most popular posts (in terms of upvotes) as the primary source for creating our conversational dataset. The selected sub-threads were as follows: \textit{r/plantcare}, \textit{r/botany}, \textit{r/houseplants}, and \textit{houzz/plantcare}. To retrieve data, we employed PRAW\footnote{\url{https://praw.readthedocs.io/en/stable/}} and Beautiful Soup\footnote{\url{https://beautiful-soup-4.readthedocs.io/en/latest/}}, utilizing the official data release keys for these forums. 
\vspace{-1em}

\vspace{-2em}
\subsection{Data Creation and Annotation}
 To begin with, given the specialized nature of the domain, we engaged with two botanists during the data creation process, accompanied by three students of botany who were also fluent in the English language and one English linguist for annotation and scalability to first curate a set of 50 conversations with each sample corresponding to a distinct post.
% To begin, we engaged the expertise of two domain experts to curate a set of 50 conversational samples, with each sample corresponding to a distinct post. 
Within each dialogue, a dyadic interaction unfolds between two entities: the `user' and the `assistant.' In this context, the term `user' denotes the individual who initiated a plant(ing)-related inquiry, while the `assistant' embodies our adept professional responsible for crafting the dialogue content by utilizing the comments of different users on a particular post.
% The assistant leverages the post and domain information to respond appropriately to the user's query.
We asked the English linguist to tag some essential semantics information about the utterances and the visual information, such as the user's intent and the corresponding dialogue acts (DAs). The intent categories are \textit{Suggestion, Conformation, Feedback}, and \textit{Awareness}, and the DA categories used are \textit{Greeting (g), Question (q), Answer (ans), Statement-Opinion (o), Statement-Non-Opinion (s),  Agreement (ag), Disagreement (dag), Acknowledge (a)} and \textit{Others (oth)}.

% \vspace{-3.5px}
In Figure \ref{dialog_creation}, we show a sample of curated conversation from our {\em Plantational} dataset. The guidelines for converting the discussion on forums into a multi-turn conversation (after consultation from the team of specialists) are detailed as follows : (i) Each discussion encompasses a question posted by the user and the subsequent comments containing relevant answers. Each of these discussions also includes an image related to the plant and its corresponding query. We convert this post (seen in Figure \ref{dialog_creation}(a)) into a multi-turn conversation between the user and an agent (seen in Figure \ref{dialog_creation}(b)) in which the user queries about its plant(ing) related issues; (ii) To achieve this, our first step entails identifying the main concern posed by the primary user sharing the content. We achieve this by analyzing both the title and the user's comments. Following this, we ascertain the most appropriate response by reviewing additional comments and selecting relevant utterances; (iii) we formulate additional questions from the post authored by the primary user (if the post encompasses multiple inquiries). 
% If a comment addresses questions pertinent to the original post, we also create questions based on that comment's content. 
The two botanists and the English linguists helped us create a set of 50 conversations from these unique posts. While creating dialogues from the primary post, the team of specialists ensured that all dialogues were coherent and prudent. Next, we focused on scaling the dataset to a reasonable size. For this, the team of student botanists was trained with 30 curated conversations to form dialogues from the raw posts. They were then presented with the remaining 20 examples to create dialogues from the raw posts. The curated dialogues by the students were then evaluated (against the gold standard) to identify their flaws, and they were again asked to correct them. Finally, the student botanists were presented with the remaining 1100 original posts and were asked to convert them into a dyadic conversation (as detailed above). We also checked the quality of these dialogues beyond the initial evaluation by randomly picking ten dialogues and checking their quality for each annotator. This process was repeated twice to ensure that the student botanists performed the task well. The two botanists conducted the quality check and also helped prepare guidelines for creating dialogues from posts. In this manner, we curated 1150 conversations from the raw posts. The dialogues created by the student botanists were interchanged amongst each other. When any of them found that a particular dialogue did not meet the quality standards, it was rejected. Around 1k conversations were included in the final corpus, which all the annotators found to be acceptable, and the remaining 150 conversations were rejected. We used Fleiss kappa \cite{fleiss1971measuring} to measure the agreement among the annotators while labeling the intents and dialogue acts. Fleiss kappa is used to measure the reliability of agreement among the annotators in categorical classification tasks. We found the Fleiss kappa score for intent labeling to be 0.72 and for dialogue acts labeling to be 0.68, which indicates good annotation quality.

\vspace{-1em}
\subsection{Plantational Dataset} 
\begin{figure*}
\centering
\subfloat[\label{intent}]{\includegraphics[width=0.39\linewidth]{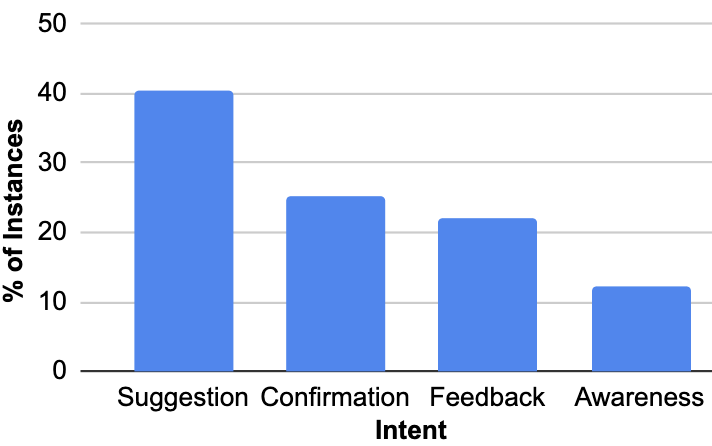}}
\subfloat[\label{DA}]{\includegraphics[width=0.36\linewidth]{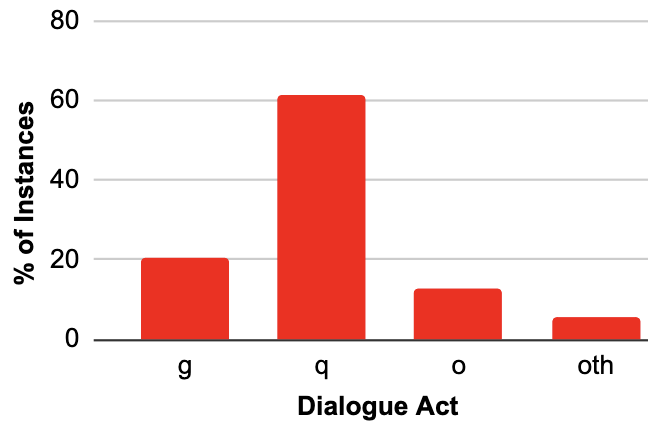}}
% \subfloat[\label{<figure4>}]{\includegraphics[width=60mm]{Figure_baseline2.png}}

\caption{Class distribution in terms of \% representation (a) for intent categories, (b) for DA categories in the {\em Plantational dataset}}
  \label{tags}
\end{figure*}

The {\em Plantational} dataset now comprises around 1K conversations between a user and a plant care assistant, amounting to over 4900 utterances. Each of these dialogues also encompasses an image of the plant and its corresponding query. The dataset also contains tags corresponding to the intent and the dialogue act for each dialogue and utterance, respectively, the distribution of which is shown in Figure \ref{tags}. Below, we study the role of incorporating multimodal features such as images with the text in each dialogue. 
\vspace{-2em}

\begin{table}[hbt!]
\centering
\caption{Statistics from the \textit{Plantational} dataset
}
\begin{tabular}{|l|r|}
\hline
\textbf{Statistics}           &  \textbf{Instances}   \\ \hline
\#Dialogues           & 963    \\ 
\#Utterances          & 4914   \\ 
\#Unique Tokens (Distinct words)
% \tablefootnote{Tokens refer to distinct words.}
& 12,953 \\ 
Average \#Utterances & 5.1    \\ 
Maximum \#Utterances & 12     \\ 
Minimum \#Utterances & 2      \\ 
% Average \#Tokens     & 2.63   \\ \hline
\#Dialogues having images     & 796   \\ \hline
\end{tabular}

\label{data_statistics}
\vspace{-2em}
\end{table}

\vspace{-1.5em}

\vspace{-1.5em}
\hspace{-0.57cm}\textbf{Role of Multimodality.} In Figure \ref{dialog_example}, we find that the user is trying to ask about the plant's identity. Also, the user is asking why the plant started dying. The model can better understand both questions if an image of the plant the user is talking about is also available. Here, we see that multimodality helps simplify generating relevant responses.
\vspace{-0.5em}

% \subsection{Ethical Consideration} 

\section{Methodology}
\begin{figure}[hbt!]
    \centering
    \includegraphics[width=\columnwidth]{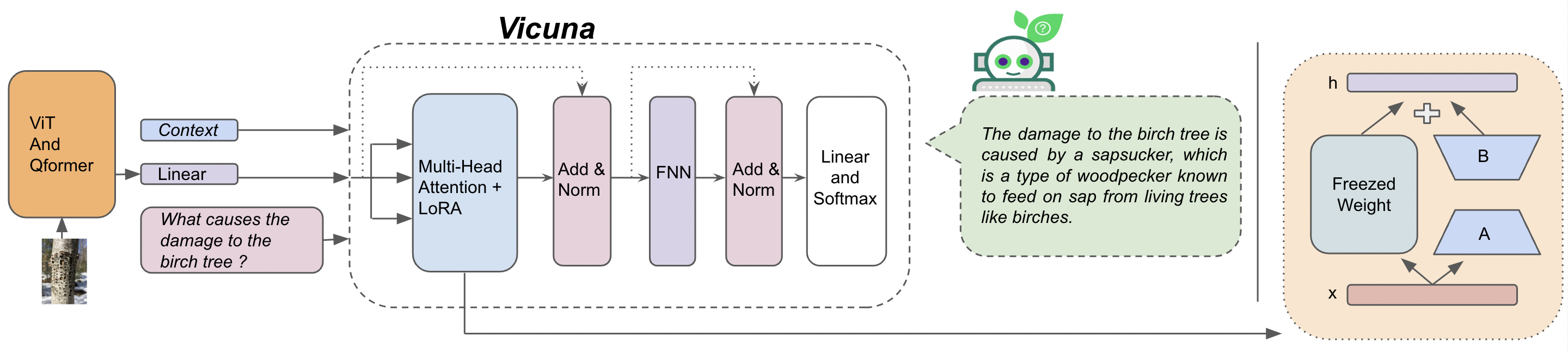}
    \caption{Proposed model. $A$ and $B$ represent the LoRA modules. Frozen weight represents the frozen weights of Multi-head attention; $x$ and $h$ are hidden representations before and after applying the LoRA module. In our model, LoRA and Linear Projection Layer are trainable while the rest is frozen. }
    \label{fig3}
\end{figure}

 The proposed methodology aims to create a plant assistant that generates contextually relevant responses to user queries. Leveraging the power of LLMs and VLM, we first analyze the efficacy of these models for plant care assistants employing various instruction-tuning (zero-shot and few-shot prompting) and fine-tuning techniques. For our proposed model, within the constraints of our small-scale dataset and recognizing the pivotal role of plant visuals in this domain, we introduce an adapter-based, fine-tuned vision-language model that effectively incorporates visual information and the textual context in its user responses. Figure \ref{fig3} illustrates this novel architecture's schematic representation.

% In this section, we describe our method. We have used various visual language models as well as text-based language models. We then fine-tune these models by using parameter-efficient fine-tuning methods. We have experimented with different models and parameter-efficient fine-tuning methods. Still, to explain the process of our experiment, we consider MiniGPT-4 \cite{zhu2023minigpt} as our visual language model and LoRA \cite{hu2021lora} as our parameter efficient fine-tuning method. 

\vspace{-2em}

\subsection{Benchmark Setup}

We first benchmark the plant care assistant using one VLM and four LLMs described below using instruction and fine-tuning strategies. 

\begin{itemize}
     \item \textbf{Llama-2} \cite{Touvron2023Llama2O} is an open-source fine-tuned language model ranging from 7 to 70 billion parameter count. It is optimized for dialogue purposes.
    \item \textbf{Vicuna} \cite{vicuna2023} is an open-source chatbot obtained by fine-tuning LLaMA on 70K user shared conversations. 
    % from ShareGPT\footnote{https://sharegpt.com/}. 
    Vicuna-13B attains 90\% quality of ChatGPT. 
    \item \textbf{FLAN-T5} \cite{chung2022scaling} is based on applying instruction fine-tuning. It studies the performance 
    of various tasks and models at different scales.
    % of instruction fine-tuning on a large number of tasks, models at different scales and on chain of thought data.
    \item \textbf{OPT} \cite{zhang2022opt} is a decoder-only model created for the purpose of matching the performance of GPT-3 \cite{brown2020language} and to study LLM performance on various tasks.
    % bring more researchers to study the performance of large language models on various tasks.
   \item \textbf{GPT-Neo} \cite{gpt-neo} It is an autoregressive language model like GPT-2 \cite{radford2019language} trained on Pile \cite{gao2020pile} dataset. It uses local attention in every other layer.

\end{itemize}
\vspace{-1.5em}

\subsection{Proposed Model}

In this section, we discuss the details of our proposed method.

\paragraph{\textbf{Textual and Visual Encoding.}} For obtaining the visual representation of the user-provided image in our proposed method, we use the vision and language representation learning part of BLIP-2 \cite{li2023blip}. It consists of ViT \cite{dosovitskiy2020image} and Q-former. The role of the ViT is to act as a vision encoder and send the visual representation to the Q-former. 
The Q-former consists of a transformer network that extracts features from the vision encoder and passes it to the projection layer. The role of the linear projection layer \cite{zhu2023minigpt} is to align the visual features with the textual features and pass the representation to the language decoder. We use Vicuna \cite{vicuna2023} as the language decoder. It is trained by instruction fine-tuning LLaMA on the conversation dataset.

\paragraph{\textbf{Parameter Efficient Fine-tuning.}}
We concatenate the image and text embedding (obtained from the user's input) and pass them to the Vicuna language model \cite{vicuna2023}. We insert LoRA blocks inside the Vicuna decoder. The LoRA module and the Linear projection layer (projecting from image space to text space) are kept trainable while the rest of the model is frozen. LoRA consists of a low-rank decomposition matrix injected into the pre-trained transformer model. It involves the following operation:- For a weight matrix, $W_o \in \mathbb{R} ^ {d \times e}$, it is updated in the following way: $W_o + \Delta W = W_o + BA$, where $B \in \mathbb{R}^{d \times f}$ and $A \in \mathbb{R}^{f \times e}$. Here, $f \ll min(d,e)$, $A$, and $B$ are trainable parameters while $W_o$ is frozen. Finally, the modified hidden representation can be described by the following equation:

\begin{equation} \label{eq:1}
h = W_ox + \Delta Wx = W_ox + BAx    
\end{equation}

Also, the $\Delta W$ matrix is scaled by a factor of $\frac{\alpha}{f}$ where $\alpha$ is a constant. 

\paragraph{\textbf{Response Generation.}}
In the final step, we take the representation coming from the lower layers of the Vicuna and transform the hidden dimension to vocabulary size using a linear projection layer. We then use a beam search algorithm to sample the following tokens and stop the generation when an end-of-sentence token is generated, or the model has generated the maximum number of tokens (defined by the user). The generated response is ideally expected to contain information about the image context and answer the user query related to the image of the plant.

\subsection{Implementation Details}

We implement the LLMs and VLM using the transformers hugging face library \cite{wolf-etal-2020-transformers} in PyTorch framework \cite{paszke2019pytorch} using a GeForce RTX 3090 GPU. LoRA is implemented using the peft library \cite{peft}. The train, validation, and test set comprises 80\%, 10\% and 10\% of the conversational instances from the {\em Plantational} dataset. The hyperparameters used are as follows: number of epochs (5), learning rate (1e-4), LoRA dropout (0.1), LoRA alpha (32), LoRA dimension (8), generated tokens (30), optimizer (Adam). We evaluate the performance of the baseline models on our {\em Plantational} dataset using automated metrics such as ROUGE \cite{lin2004rouge}, BLEU \cite{papineni2002bleu} and BERT scores \cite{zhang2019bertscore}.
We also perform human evaluation on metrics such as
% We as well evaluate the quality of the response generated using several human evaluation metrics such as 
\textbf{(i)} \textit{Fluency}: The response must be grammatically correct; \textbf{(ii)} \textit{Adequacy}: To generate a response related to user's query; \textbf{(iii)} \textit{Informativeness}: To generate the response that answers user's problem; \textbf{(iv)} \textit{Contextual Relevance}: To generate response related to the context of the conversation; \textbf{(v)} \textit{Image Relevance}: To generate response related to the image. We performed human evaluation across 95 dialogues using three human evaluators (from the authors' affiliation) and reported the average score.

\section{Results and Discussion}
To assess the performance of 
various dialogue generation models for plant care, we employed commonly used evaluation metrics, including BLEU and ROUGE. We also assessed the effectiveness of various models in terms of the semantic alignment of their generated text with the reference (gold) responses. Moreover, we conducted a human evaluation to mitigate the risk of under-assessment performed by automatic evaluation metrics. Our discussion commences with an examination of the experimental results we have obtained. Subsequently, we delve into the findings and evidence pertaining to the research questions, culminating in the presentation of a case study showcasing the performances of various models.

\vspace{-1.5em}
\begin{table}[hbt!]
\centering
\caption{Performances of different models for plant assistance response generation}
\resizebox{0.9\linewidth}{!}{
\begin{tabular}{|cc|ccc|cccc|c|c|}
\hline
\multicolumn{2}{|c|}{\multirow{2}{*}{\textbf{Model}}}                                                          & \multicolumn{3}{c|}{\textbf{ROUGE}} & \multicolumn{4}{c|}{\textbf{BLEU}}    & \multirow{2}{*}{\textbf{BLEU}} & \multirow{2}{*}{\textbf{BERT-F1}} \\ \cline{3-9}
\multicolumn{2}{|c|}{}                                                                                & R-1       & R-2     & R-L     & B1    & B2    & B3    & B4   &                          &                          \\ \hline
\multicolumn{1}{|c|}{\multirow{3}{*}{Flan}}                                               & zero-shot & 18.94    & 6.27   & 16.64  & 8.01  & 3.41  & 1.95  & 0.36 & 3.43                     & 57.98                    \\
\multicolumn{1}{|c|}{}                                                                    & few-shot  & 22.29    & 7.46   & 19.86  & 9.25  & 4.75  & 2.42  & 0.70 & 4.28                     & 59.71                    \\
\multicolumn{1}{|c|}{}                                                                    & \textbf{fine-tune} & \textbf{29.80}    & \textbf{14.61}  & \textbf{26.63}  & \textbf{16.43} & \textbf{11.12} & \textbf{8.58}  & \textbf{3.91} & \textbf{10.01}                    & \textbf{63.42}                    \\ \hline
\multicolumn{1}{|c|}{\multirow{3}{*}{\begin{tabular}[c]{@{}c@{}}GPT-Neo\end{tabular}}} & zero-shot & 14.702   & 3.45   & 11.90  & 6.71  & 2.19  & 0.95  & 0.35 & 2.55                     & 50.15                    \\
\multicolumn{1}{|c|}{}                                                                    & few-shot  & 17.13    & 5.17   & 14.21  & 8.37  & 3.30  & 1.60  & 0.73 & 3.5                      & 51.25                    \\
\multicolumn{1}{|c|}{}                                                                    & \textbf{fine-tune} & \textbf{30.39}    & \textbf{15.09}  & \textbf{26.94}  & \textbf{18.67} & \textbf{12.10} & \textbf{9.08}  & \textbf{6.30} & \textbf{11.53}                    & \textbf{59.59}                    \\ \hline
\multicolumn{1}{|c|}{\multirow{3}{*}{OPT}}                                                & zero-shot & 17.80    & 4.79   & 14.87  & 7.96  & 2.81  & 1.29  & 0.51 & 3.14                     & 50.69                    \\
\multicolumn{1}{|c|}{}                                                                    & few-shot  & 20.90    & 6.65   & 17.70  & 10.49 & 4.21  & 1.80  & 0.96 & 4.36                     & 52.72                    \\
\multicolumn{1}{|c|}{}                                                                    & \textbf{fine-tune} & \textbf{29.07}    & \textbf{16.44}  & \textbf{26.27}  & \textbf{18.64} & \textbf{13.47} & \textbf{10.39} & \textbf{6.33} & \textbf{12.20}                    & \textbf{59.59}                    \\ \hline

\multicolumn{1}{|c|}{\multirow{3}{*}{Llama-2}}                                                & zero-shot & 19.47    & 4.83   & 15.57  & 8.55  & 3.12  & 1.50  & 0.78 & 3.48                   &  57.20                    \\
\multicolumn{1}{|c|}{}                                                                    & few-shot  & 21.96    &  6.62   & 17.64  &  10.14 & 4.42  & 2.45  & 1.48 & 4.62                   &  58.93                    \\
\multicolumn{1}{|c|}{}                                                                    & \textbf{fine-tune} & \textbf{30.30}    & \textbf{15.74}  & \textbf{27.19}  & \textbf{17.79} & \textbf{11.86} & \textbf{8.62} & \textbf{5.26} & \textbf{10.88}                    & \textbf{60.86}                    \\ \hline

\multicolumn{1}{|c|}{\multirow{3}{*}{Vicuna}}                                             & zero-shot & 19.84    & 5.29   & 16.28  & 9.03  & 3.59  & 1.77  & 0.97 & 3.84                     & 55.49                    \\
\multicolumn{1}{|c|}{}                                                                    & few-shot  & 20.33    & 5.09   & 16.46  & 8.81  & 3.11  & 1.38  & 0.81 & 3.52                     & 55.26                    \\
\multicolumn{1}{|c|}{}                                                                    & \textbf{fine-tune} & \textbf{30.34}    & \textbf{16.24}  & \textbf{27.20}  & \textbf{18.35} & \textbf{12.66} & \textbf{9.98}  & \textbf{6.49} & \textbf{11.87}                    & \textbf{61.13}                    \\ \hline
\multicolumn{1}{|c|}{\multirow{3}{*}{\textbf{\em EcoSage}}}                                               & zero-shot & 22.16    & 6.49   & 18.13  & 10.32 & 4.45  & 2.59  & 1.72 & 4.77                     & 56.64                    \\
\multicolumn{1}{|c|}{}                                                                    & few-shot  & 19.97    & 5.83   & 16.52  & 9.00  & 3.86  & 2.21  & 1.47 & 4.13                     & 54.30                    \\
\multicolumn{1}{|c|}{}                                                                    & \textbf{fine-tune} & \textbf{25.35}    & \textbf{11.76}  & \textbf{21.77}  & \textbf{14.07} & \textbf{8.37}  & \textbf{5.65}  & \textbf{3.95} & \textbf{8.01}                     & \textbf{58.76}                    \\ \hline
\end{tabular}
}
\label{baseline_table}
\end{table}
\vspace{-1.5em}

\subsection{Experimental Results} Table \ref{baseline_table} summarizes the performances of different LLMs and VLM for appropriate plant assistance response generation. We have reported results for three different settings, namely zero-shot, few-shot, and fine-tuning. Across various  LLMs such as Flan, GPT-Neo, OPT, and Vicuna, we observe a consistent upward trend in performance when transitioning from zero-shot to few-shot to fine-tuning settings. In a few settings, particularly Vicunna and the proposed model, the performance of zero-shot has been superior to few-shot. The most probable reason seems to be the generalizability of these two highly large models with the handful number of samples. We found an unusual finding that the textual  LLM performs better than the  VLM within the context of {\em EcoSage}, despite {\em EcoSage} including both user query text and images related to plants. The decrease in performance happens primarily due to the challenge of aligning the visual embedding with the text embedding. Integrating visual information alongside textual data may introduce complexities in the model's embedding, resulting in a less effective alignment and leading to poorer overall performance when compared to LLMs. 

We also carried out an ablation study involving different settings. The obtained result has been reported in Table \ref{ablation_minigpt4}. We experimented with another visual encoding method called Data2vec \cite{baevski2022data2vec}; it did not outperform the uni model. We also examined various image configurations: textual dialogues with a blank white image, textual dialogues with images (whenever available), and textual dialogues with images (whenever available), along with a blank white image for dialogues that do not include visual descriptions. The results indicate that adding visual descriptions does improve the model's ability to offer suitable suggestions. It faces challenges in aligning visual and textual embedding space. 
\vspace{-1.5em}

\begin{table}[hbt!]
\centering
\caption{Performance of {\em EcoSage} with different modalities}
\resizebox{0.6\linewidth}{!}{
\begin{tabular}{|l|c|c|c|}
\hline

\textbf{Model} & \textbf{ROUGE-L} & \textbf{BLEU-4} & \textbf{BERT-F1} \\ \hline     
VE (Data2vec) & 20.86   & 3.33   & 56.85   \\
T + I (all blank)  & 21.40   & 3.86   & 57.21   \\ 
T  + I (available)     & 21.77   & 3.95   & 58.76   \\ 
T + I    & \textbf{22.82}   & \textbf{4.27}   & \textbf{55.90}   \\ \hline
\end{tabular}
}
\label{ablation_minigpt4}
\end{table}
\vspace{-1.5em}

\begin{figure}[ht]
    \centering
    \includegraphics[width=0.8\linewidth]{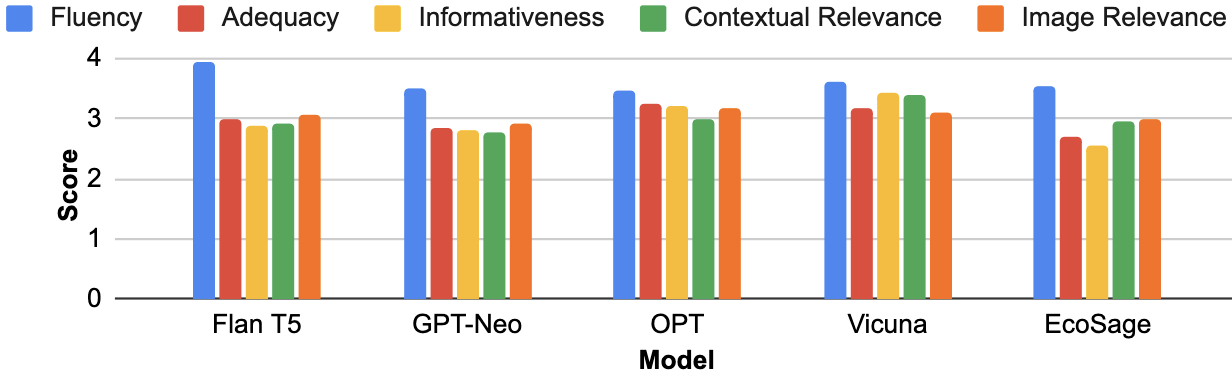}
    \caption{Human evaluation scores of different models based on diverse metrics}
    \label{human_eval}
\end{figure}

\paragraph{\textbf{Human Evaluation.}}
Figure \ref{human_eval} presents results based on a human evaluation study. 
We have observed that all the models excel in terms of fluency. However, both LLMs and VLM face challenges regarding adequacy, informativeness, contextual relevance, and image relevance. 
This suboptimal performance of LLMs and VLM can be attributed to the specific and detailed nature of user queries related to plants. 
% Even though these models have undergone fine-tuning using our dataset, they struggle due to a lack of specialized knowledge in the domain. 
Furthermore, it is noteworthy that the VLM exhibits a poorer performance compared to LLMs. It can be explained due to the VLM's inability to understand images and can be enhanced by training on image-text pairs.
% It can be explained as VLM inability to understand image and its performance can be improved by
% This discrepancy can be attributed to VLMs' limitations in comprehending plant-related images. VLM performance can be enhanced through 
% additional training using image-text pairs related to plants.

\begin{figure}[hbt!]
    \centering
    \includegraphics[width=0.85\columnwidth]{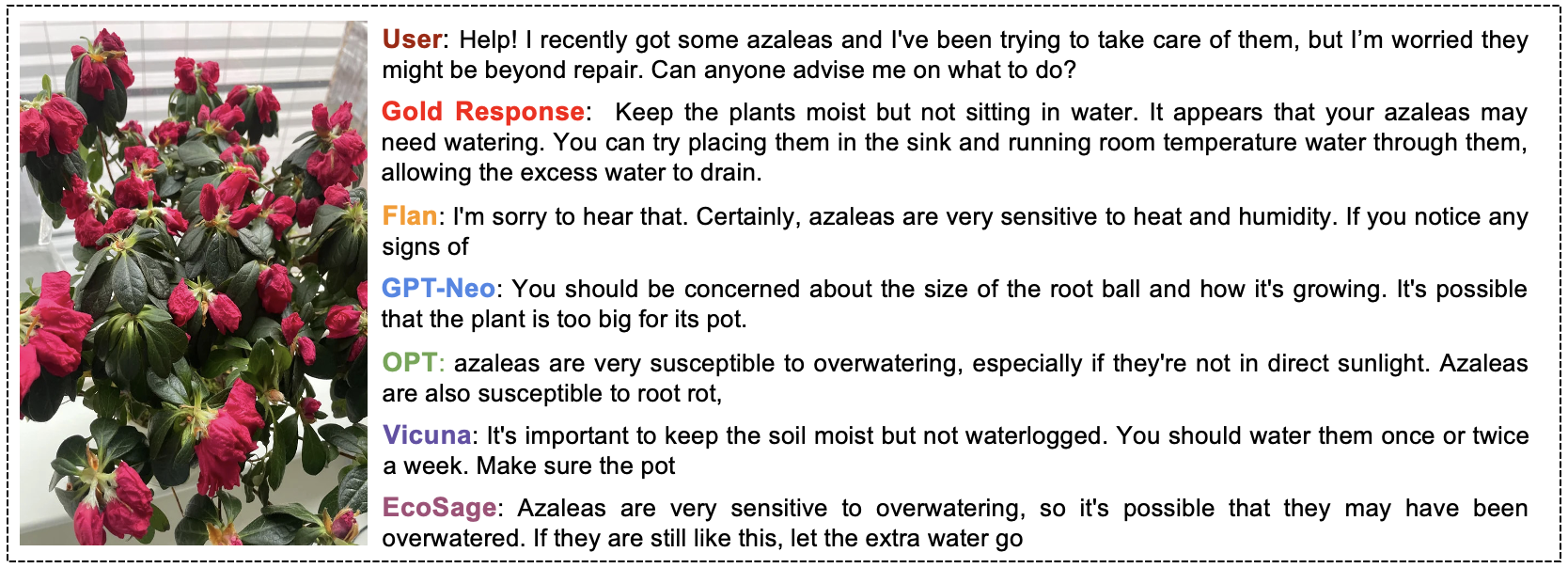}
    \caption{Qualitative analysis of response generated by different models.}
    \label{fig:qualitative_example}
\end{figure}

\paragraph{\textbf{Qualitative Analysis.}}
We examined the text generation capabilities of various models using a representative example from the {\em Plantational} dataset, as depicted in Figure \ref{fig:qualitative_example}. Our analysis reveals that Vicuna produced a response that closely aligns with the gold standard response. It is followed by responses from {\em EcoSage}, OPT, Flan, and GPT-Neo, in descending order of similarity to the gold response.

\vspace{-1em}
\vspace{-1.5em}

\subsection{Findings to Research Questions} Based on the experiments, we report the following answers (with evidence) to our investigated research questions (RQs). \\

\hspace{-0.57cm}\textbf{RQ1:  Can existing state-of-the-art LLMs adequately offer initial recommendations related to plant(-ing)-related queries?} Based on results from Table \ref{baseline_table} and Table \ref{human_eval} (zero-shot and few-shot), it becomes evident that LLMs face challenges in delivering satisfactory responses to user queries pertaining to the plantation. This underscores the necessity for LLMs to acquire more domain-specific knowledge about plants and improve their ability to contextualize information from images. We observed a noteworthy enhancement, varying from 10\% to 50\%, when employing both few-shot learning and fine-tuning across various evaluation metrics.

\hspace{-0.55cm}\textbf{RQ2: Do LLMs that take images into account comprehend concerns more effectively and produce suitable and better responses? } The results we obtained (as shown in Table \ref{ablation_minigpt4}) provide affirmative evidence supporting the assertion. The model that considers images along with textual description has obtained superior performance and generates a more context-specific response.

\hspace{-0.55cm}\textbf{RQ3: What is the appropriate way to gauge the effectiveness of the response generation model? Are metrics based on n-gram overlap sufficient for evaluation, and are they consistent with semantic evaluation?} In numerous instances, we noticed that the generated response is highly contextually relevant but exhibits limited overlap with the reference response (as shown in Figure \ref{fig:qualitative_example}), resulting in lower evaluation scores such as BLEU and ROUGE. Therefore, we also incorporated the measurement and reporting of BERT-F1, a semantic-based evaluation metric that assesses the similarity between the semantic embeddings of the reference response and the generated sentence. Consequently, the Vicunna model achieves the highest BERT-F1 score despite having a lower BLEU score. Henceforth, we firmly support that a comprehensive evaluation of effectiveness should encompass both lexical and semantic-based assessments.

\section{Conclusion} In this paper, we make the first move to investigate some fundamental research questions related to plant assistance response generation and build an assistant called {\em EcoSage} to provide guidance to plant(-ing)-related issues of the users. Confronted by the scarcity of conversational data in plant care, we undertook the initiative to create a plant care conversational dataset named {\em Plantational} encompassing a range of plant(-ing)-related issues. We further evaluate LLMs and VLM on the {\em Plantational} dataset by generating responses corresponding to the user's query in zero-shot, few-shot, and fine-tune settings. Additionally, our proposed \textit{EcoSage} is a multi-modal model, a plant assistant utilizing LoRA units for adapting it to plant-based conversations. In the future, we aim to focus on improving the image understanding capabilities of VLM for it to perform better in generating coherent responses than LLMs.

\section{Acknowledgement}
Dr. Sriparna Saha extends heartfelt gratitude for the Young Faculty Research Fellowship (YFRF) Award, supported by the Visvesvaraya Ph.D. Scheme for Electronics and IT, Ministry of Electronics. Abhisek Tiwari expresses sincere gratitude for the support received by the Prime Minister Research Fellowship (PMRF) Award provided by the Government of India. This grant has played a role in supporting this research endeavor. We also thank Rachit Ranjan, Ujjwal Kumar, Kushagra Shree, and other annotators for the dataset development.

% \bibliography{Ref}

\end{document}